\pdfoutput=1
\documentclass[11pt]{article}

\usepackage[final]{acl}

\usepackage{times}
\usepackage{latexsym}

\usepackage[T1]{fontenc}
\usepackage[utf8]{inputenc}

\usepackage{microtype}

\usepackage{inconsolata}

\usepackage{graphicx}

\usepackage{multirow}%
\usepackage{amsmath,amssymb,amsfonts}%
\usepackage{algorithm}%
\usepackage{algpseudocode}%
\usepackage{booktabs}

\title{DipSVD: Dual-importance Protected SVD for Efficient LLM Compression}

\author{
 \textbf{Xuan Ding\textsuperscript{1}},
 \textbf{Rui Sun\textsuperscript{2}},
 \textbf{Yunjian Zhang\textsuperscript{4}},
 \textbf{Xiu Yan\textsuperscript{5}},
 \textbf{Yueqi Zhou\textsuperscript{1}},
\\
 \textbf{Kaihao Huang\textsuperscript{1}},
 \textbf{Suzhong Fu\textsuperscript{2}},
 \textbf{Chuanlong Xie \textsuperscript{1}},
 \textbf{Yao Zhu\textsuperscript{3}}
\\
\\
 \textsuperscript{1}Beijing Normal University, \\
 \textsuperscript{2}The Chinese University of HongKong, Shenzhen,
 \textsuperscript{3}Zhejiang University, \\
 \textsuperscript{4}University of Chinese Academy of Sciences,
 \textsuperscript{5}Tsinghua University
\\
 \small{
   202322011119@mail.bnu.edu.cn, ee\_zhuy@zju.edu.cn
 }
}

\begin{document}
\maketitle
\begin{abstract}
The ever-increasing computational demands and deployment costs of large language models (LLMs) have spurred numerous compressing methods. Compared to quantization and unstructured pruning, SVD compression offers superior hardware compatibility and theoretical guarantees. However, existing SVD-based methods focus on the overall discrepancy between the original and compressed matrices while overlooking the protection of critical components within the matrix, which leads to inferior performance in the compressed models. This paper proposes a dual-level importance protection mechanism to enhance SVD-based compression methods: (1) local importance protection: preserving the most critical singular vectors within each weight matrix through channel-weighted data whitening; and (2) global importance protection: enabling less important layers to bear a greater portion of the compression burden through either a heuristic or optimization-based approach, thereby minimizing the impact of compression on critical layers. Extensive experiments demonstrate that DipSVD outperforms existing SVD-based compression approaches across multiple benchmarks, achieving superior model performance especially at high model compression ratios.
\end{abstract}

\section{Introduction}

While Large Language Models (LLMs) demonstrate remarkable capabilities across diverse natural language tasks such as multi-round conversation \citep{chen2023chatcot, long2023large} and logical reasoning \citep{creswell2022selection, duan2024gtbench, pan2023logic},  the ever-increasing model scales impose severe computational burdens \citep{zhou2024survey, wang2024iot}.  This has spurred intensive research into LLM-specific compression techniques, including quantization \citep{frantar2022gptq,lin2024awq}, parameter pruning \citep{men2024shortgpt, ma2023llm, kim2024shortened, song2024sleb,ding2025sliding}, and knowledge distillation \citep{gu2023minillm,hsieh2023distilling}. While quantization and pruning require specialized hardware support and costly retraining, low-rank decomposition methods like Singular Value Decomposition (SVD) offer hardware-agnostic compression through dense matrix operations. Moreover, the KV cache of LLMs compressed via SVD at runtime can also be reduced \citep{wang2024svd}.

\begin{table}[t]
\centering
\fontsize{60pt}{20pt}\selectfont 
\renewcommand{\arraystretch}{5.0}
\resizebox{1.0\columnwidth}{!}{%
\begin{tabular}{c|lll}
\hline \hline
\ \textbf{Method}& \ Local Importance& \ Global Importance& \ Coupling Modeling\\ \hline
\ FWSVD& \ $\checkmark$  Row-wise Fisher weighting& \ \text{\sffamily X}& \ \text{\sffamily X}\\
\ ASVD& \ \text{\sffamily X}& \ $\checkmark$ Sensity-based Truncation Rank Searching& \ \text{\sffamily X}\\
\ SVD-LLM& \ \text{\sffamily X}& \ \text{\sffamily X}& \ \text{\sffamily X}\\
\ Ours& \ $\checkmark$ Channel-weighted Data
Whitening& \ $\checkmark$ Layer-Specific Compression& \ $\checkmark$ Cross-Hierarchy Joint Optimization\\ \hline\hline
\end{tabular}}
\caption{Comparison of existing SVD-based methods.}
\label{motivation}
\end{table}

Despite these advantages, existing SVD-based compression methods are undermined by their failure to holistically consider both \textbf{local importance} (e.g., intra-layer channel sensitivity) and \textbf{global importance} (e.g., layer-wise heterogeneity) during matrix factorization (as shown in Tab.\ref{motivation}):  (1) Globally-aware methods like ASVD \citep{yuan2023asvd} dynamically allocate compression ratios across layers but retain standard SVD decomposition within each layer, risking excessive pruning of sensitive local features. (2) Locally-aware methods such as FWSVD \citep{hsu2022language} weight intra-layer channels but ignore global disparities. In LLMs, where layers with same structure have different roles \citep{zhang2024investigating}, uniform compression across these heterogeneous layers leads to suboptimal efficiency. (3) Isotropic methods including SVD-LLM \citep{wang2024svd} apply homogeneous compression without importance weighting at either level and inevitably degrade the performance-efficiency trade-offs. Critically, no existing work jointly optimizes global and local importance during SVD decomposition—a gap we empirically prove to be detrimental under aggressive compression.

In this paper, we propose DipSVD, a \textbf{D}ual-\textbf{i}mportance \textbf{p}rotected SVD-based compression method. Specifically, our approach introduces: (1) Local Importance Protection: By employing channel-weighted data whitening, this method preserves the most importance singular vectors in the weight matrix, while allowing less important singular vectors to bear a greater portion of the compression burden. (2) Global Importance Protection: Through either an optimization-based approach or a heuristic method derived from layer-wise gradient sensitivity analysis, we automatically determine the optimal layer-specific compression ratios, thereby protecting the most critical layers of the model and allocating more compression burden to less importance layers. Through Pearson correlation analysis, we verify that the heuristic compression ratios closely align with those obtained from Bayesian optimization, while simultaneously reducing computation overhead. 

We conduct extensive experiments to evaluate the effectiveness of our DipSVD method, benchmarking against three SVD-based compression methods across five LLMs of varying architectures and scales. Our evaluation encompasses both zero-shot task performance and perplexity metrics under identical experimental conditions. Experimental results demonstrate the superiority of DipSVD in terms of both zero-shot task performance and generation quality. Additionally, our ablation studies show that with one of the two key components of DipSVD alone, it still outperforms state-of-the-art SVD-based compression methods under different compression ratios.

The contributions of this study are summarized as:
\begin{itemize}
\item We propose DipSVD, a novel compression framework for LLMs that explicitly integrates both local and global importance into the SVD process, aiming to preserve model integrity and task performance.
\item To implement this framework, we (i) introduce an importance-aware whitening mechanism to efficiently estimate compression loss while emphasizing local importance, and (ii) develop two strategies for global importance modeling: a high-performance Bayesian optimization method and a lightweight heuristic-based alternative.

\item Extensive experiments demonstrate that DipSVD effectively improves on-device inference efficiency while maintaining superior performance across seven diverse zero-shot tasks and perplexity benchmarks.

\end{itemize}

\begin{figure*}[h]
\centering
\includegraphics[width=0.95\textwidth]{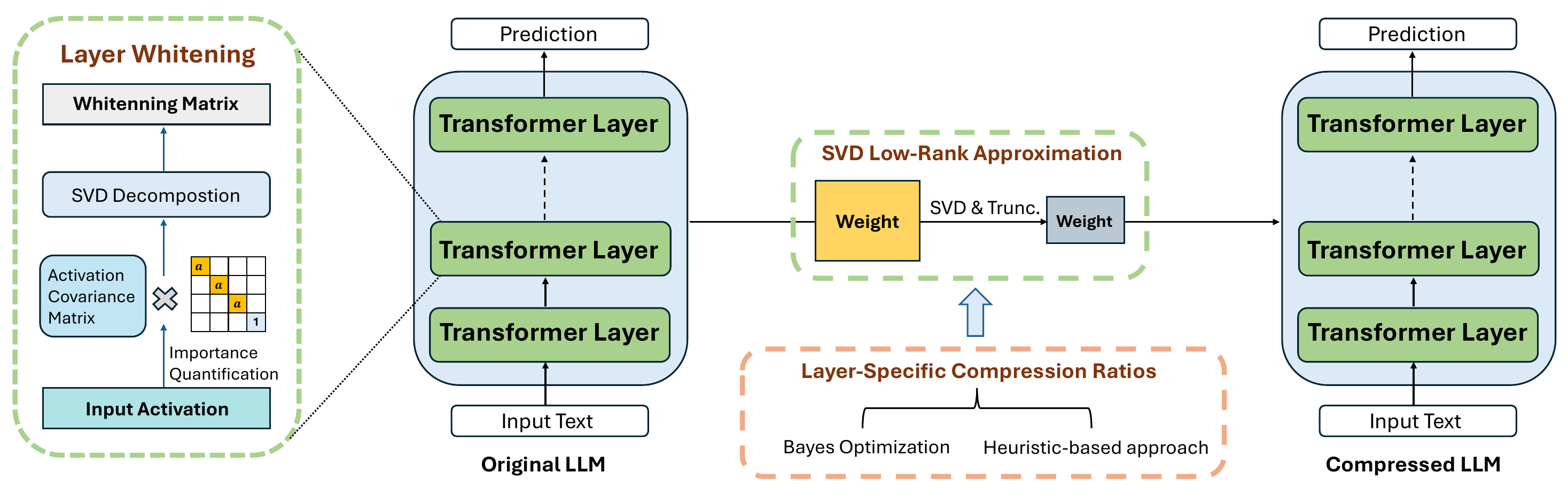}
\caption{Overview of DipSVD.}
\label{framework}
\end{figure*}

\section{Related Work}
\subsection{Large Language Model  Compression}
To mitigate the computational and memory demands of LLMs, researchers have proposed multiple compression techniques such as quantization, unstructured pruning, structured pruning  and knowledge distillation. Although these methods have demonstrated practical effectiveness, each has its own limitations. Quantized methods like GPTQ \citep{frantar2022gptq}, AWQ \citep{lin2024awq}, and SmoothQuant \citep{xiao2023smoothquant} enable low-bit inference but often sacrifice accuracy or hardware efficiency. Unstructured pruning methods like SparseGPT \citep{frantar2023sparsegpt}, Wanda \citep{sun2023simple} and GBLM-Pruner \citep{das2023beyond} can remove 50–75\% of weights with reasonable accuracy, though their practical speedups on general-purpose hardware remain limited. While structured pruning methods \citep{chen2021actnn,ma2023llm,an2024fluctuation} achieve hardware acceleration through removing entire architectural components, they suffer from substantial accuracy loss under aggressive pruning. Recent knowledge distillation methods such as GKD \citep{agarwal2023gkd} and DistiLLM \citep{ko2024distillm} compress auto-regressive models into smaller students, but require substantial computation and retraining. Unlike these methods, low-rank approximation based on Singular Value Decomposition (SVD) offers an efficient alternative, requiring no retraining while maintaining hardware compatibility.

\subsection{Low-rank Decomposition}
SVD compression is a widely used technique to reduce matrix size by decomposing the weight matrix and truncating smaller singular values \citep{kim2015compression}. Specifically, FWSVD \citep{hsu2022language} incorporates Fisher information to prioritize important parameters, better maintaining prediction accuracy.  ASVD \citep{yuan2023asvd} scales the weight matrix based on activation distributions and adjusts the compression ratio layer-wise, allowing performance preservation at moderate compression ratios without retraining.
SVD-LLM \citep{wang2024svd}  introduces truncation-aware data whitening and closed-form layer-wise parameter updates, significantly improving the balance between compression efficacy and inference speed.

However, existing SVD-based methods typically neglect both global and local importance considerations in model architecture. The failure to account for global importance, as evidenced by varying layer-wise compression sensitivity, leads to suboptimal rank selection and significant performance degradation under high compression ratios \citep{ding2025sliding}. Concurrently, ignoring local importance, manifested through unequal channel contributions to model outputs, may inadvertently prune structurally critical channels and cause substantial accuracy deterioration.

\section{Method}
Fig.\ref{framework} provides an overview of DipSVD. Following the standard procedure of post-training LLM compression methods, DipSVD first uses a random set of sentences as calibration data to generate activation for local importance protection and layer whitening. Specifically, DipSVD selectively emphasize important channels in the whitening transformation, which not only ensures a direct mapping between singular values and compression loss, but also improves structural retention during compression. To preserve global importance of LLMs, DipSVD groups the weight matrices across by Transformer layer in the original LLM. For each Transformer layer, DipSVD assigns a unique compression ratio which computed through Bayes Optimization or Heuristic-based approach. Lastly, DipSVD applies the layer-specific compression ratios to the weight matrix and performs SVD to truncate the weight matrices to compress the LLM. The following subsections provide a detailed description of each protection method and their synergistic integration. Pseudocode is provided in Appendix.A.

\subsection{Local Importance Protection}

\subsubsection{Channel-weighted Whitening}

To preserve structurally important channels during whitening, we propose an importance-aware transformation that adapts to the second-order statistics of the input. Let \(X \in \mathbb{R}^{m \times n}\) denote a data matrix with \(m\) samples and \(n\) feature channels.

We assess the structural importance of each channel by evaluating how much it contributes to the overall sample-level second-order structure. Specifically, for the \(j\)-th feature channel \(x_j = X_{:,j} \in \mathbb{R}^m\), we define its importance as:
\begin{equation}
\alpha_j = \sqrt{x_j^{T} (X X^{T}) x_j}.
\end{equation}
Channels with larger \(\alpha_j\) values are considered more significant, as they exert a stronger influence on the global sample structure encoded in \(X X^T\). 
Geometrically, this expression can be viewed as the magnitude of the projection of channel vector \(x_j\) onto the principal subspace spanned by the data samples. In other words, \(\alpha_j^2\) reflects how well the direction \(x_j\) aligns with the dominant variance structure in the sample space. Channels with high alignment are thus deemed structurally important and are prioritized for preservation in the subsequent whitening step.

To enhance the preservation of such channels during whitening, we introduce a diagonal scaling matrix \(D \in \mathbb{R}^{n \times n}\), defined by:
\begin{equation}
D_{jj} = 
\begin{cases}
a, & \text{if } \alpha_j \text{ is among the top } p\% \text{ values}, \\
1, & \text{otherwise},
\end{cases}
\end{equation}
with $a > 1$.
This results in a reweighted input \(\tilde{X} = X D\), where structurally important channels are selectively amplified.

As a preparatory step for whitening, we compute the second-order structure of \(\tilde{X}\) through the matrix product:
\begin{equation}
\tilde{X}^T \tilde{X} = D^T X^T X D.
\end{equation}
Applying singular value decomposition, we express this matrix as \(\tilde{X}^T \tilde{X} = U_{\tilde{X}} \Sigma_{\tilde{X}} U_{\tilde{X}}^T\), where \(U_{\tilde{X}} \in \mathbb{R}^{n \times n}\) is orthogonal and \(\Sigma_{\tilde{X}}\) is diagonal with non-negative entries. We then construct the whitening matrix:
\begin{equation}
S = \Sigma_{\tilde{X}}^{-1/2} U_{\tilde{X}}^T,
\end{equation}
which yields the final whitened output as \(\hat{X} = \tilde{X} S = X D S\). This process results in decorrelated, variance-normalized features, while maintaining the structural contributions of high-importance channels identified through the original covariance.

\subsubsection{Impact of Whitening on Truncation}

The channel-weighted whitening operation amplifies the contributions of important channels while ensuring that the compression loss is directly related to the singular values of the weight matrix, which is critical for minimizing the performance degradation caused by compression. In the following, we provide a theoretical derivation explaining why the whitening process guarantees a direct mapping between singular values and compression loss.

\textbf{Singular Value Decomposition of Whitened Weight Matrix.} We first perform SVD on the whitened weight matrix \( W S \) to obtain its decomposition:
\begin{equation}
W S = U \Sigma V^T = \sum_{i=1}^r \sigma_i u_i v_i^T,
\end{equation}

where: \( U = [u_1, u_2, ..., u_r] \) contains the left singular vectors \( u_i \in \mathbb{R}^n \), \( \Sigma = \text{diag}(\sigma_1, \sigma_2, ..., \sigma_r) \) contains the singular values in descending order, \( V = [v_1, v_2, ..., v_r] \) contains the right singular vectors \( v_i \in \mathbb{R}^m \), \( r \) is the rank of \( W S \).

\textbf{Single Singular Value Truncation.} When truncating the \( i \)-th singular value \( \sigma_i \) of \( W S \), the compression loss \( L_i \) is given by:
\begin{equation}
\begin{aligned}
L_i &= \|W \tilde{X} - W' \tilde{X}\|_F \\
    &= \|(W S - \text{SVD}(W S)) S^{-1} \tilde{X}\|_F \\
    &= \|\sigma_i u_i v_i^T S^{-1} \tilde{X}\|_F.
\end{aligned}
\end{equation}

Leveraging the orthonormality of \( u_i \) and \( v_i \) (i.e., \( u_i^T u_i = v_i^T v_i = 1 \)), the whitening property \( S^{-1} \tilde{X} \tilde{X}^T (S^{-1})^T = I \), and the invariance of the Frobenius norm under orthogonal transformations, we obtain: 
\begin{small} 
\begin{equation}
\begin{aligned}
L_i &= \sigma_i \left[\text{trace}\left(v_i^T S^{-1} \tilde{X} \tilde{X}^T (S^{-1})^T v_i\right)\right]^{1/2} \\
&= \sigma_i \left[v_i^T \cdot I \cdot v_i\right]^{1/2} \\
&= \sigma_i.
\end{aligned}
\end{equation}
\end{small} 

This shows that truncating a single singular value results in a compression loss equal to that singular value.

\textbf{Multiple Singular Value Truncation.}
When truncating the smallest \( r - m \) singular values \( \{\sigma_{m+1}, \dots, \sigma_r\} \) of \( W S \), the total compression loss is defined as the output difference on input \( \tilde{X} \):
\begin{small} 
\begin{equation}
L = \left\| \sum_{i=m+1}^r \sigma_i u_i v_i^T S^{-1} \tilde{X} \right\|_F 
  = \sqrt{ \sum_{i=m+1}^r \sigma_i^2 }.
\end{equation}
\end{small} 
% This result follows from three key properties:
% 1) The orthonormality of singular vectors.
% 2) The whitening condition \( S^{-1} \tilde{X} \tilde{X}^T (S^{-1})^T = I \).
% 3) The Frobenius norm's preservation under orthogonal transformations.

In summary, truncating smaller singular values minimizes both the compression loss and approximation error, where the total impact is determined by the root-sum-square of the truncated singular values. The whitening operation ensures this direct correspondence between singular values and model performance.

\subsection{Global Importance Protection}
While local importance protection effectively preserves critical channels within individual layers, a key challenge lies in ensuring that globally important layers are also adequately protected. To this end, we propose two compression strategies based on global layer-wise importance, which allocate different compression ratios to different layers according to their relative significance.

\subsubsection{Bayesian Optimization for Layer-Specific Compression}

For scenarios where computational resources are abundant and the highest possible performance is desired, Bayesian optimization can be employed to directly optimize the global compression objective. This approach searches for the optimal compression ratios by maximizing the cosine similarity between the outputs of the original and compressed models, subject to the global compression constraint: 
\begin{equation}
\begin{small} 
\begin{aligned}
\max_{k_1, k_2,\ldots,k_L} &\text{cosine\_similarity}\big(f_{\text{orig}}(x), f_{\text{comp}}(x)\big) \\
&\text{s.t. } \frac{1}{L}\sum_{l=1}^L k_l = k,
\end{aligned}
\end{small}
\end{equation}
where $k_l$ are layer compression ratios, $L$ is total layers, and $k$ is the target global compression ratio.

\subsubsection{Efficient Heuristic for Layer-Specific Compression}
While Bayesian optimization offers superior performance, it is computationally expensive and may not be necessary for all applications. A practical and efficient alternative is provided by a heuristic-based approach that combines two key metrics: Fisher Sensitivity and Effective Rank. These metrics provide complementary insights into the importance and compressibility of each layer, enabling a cost-effective yet principled compression strategy.

\textbf{Fisher Sensitivity.} Fisher Sensitivity measures how sensitive the model's loss is to changes in the parameters of each layer. It is computed as the ratio of the gradient norm to the parameter norm for each layer, capturing the relative importance of the layer's parameters. Specifically, for each layer \(l\), Fisher Sensitivity can be formulated as:
\begin{equation}
S_l = \sum_{\text{Attention}} \frac{\|\nabla_{\theta} \mathcal{L}\|_F}{\|\theta\|_F} + \sum_{\text{MLP}} \frac{\|\nabla_{\theta} \mathcal{L}\|_F}{\|\theta\|_F},
\end{equation}
where \(\nabla_{\theta} \mathcal{L}\) is the gradient of the loss with respect to the parameters \(\theta\) and \(\|\cdot\|_F\) denotes the Frobenius norm.

\textbf{Effective Rank.} Effective Rank quantifies the information density of each layer's output by analyzing its singular value distribution. For each layer \(l\), we compute the singular values of its output matrix \(H_l \in \mathbb{R}^{B \times T \times D}\) (where \(B\) is the batch size, \(T\) is the sequence length, and \(D\) is the hidden dimension) and determine the smallest rank \(R_l\) that captures a predefined threshold (e.g., 95\%) of the cumulative energy:
\begin{equation}
R_l = \min\left\{k \mid \frac{\sum_{i=1}^k \sigma_i}{\sum_{i=1}^r \sigma_i} \geq \text{threshold}\right\},
\end{equation}
where \(\sigma_i\) are the singular values of \(H_l\).

% \textbf{Combining Sensitivity and Effective Rank.} 
% Layers with higher sensitivity values are more critical to the model's performance and should be compressed less aggressively.  Similarly, layers with lower effective ranks are more compressible and can tolerate higher compression ratios. To assign compression ratios to each layer, we first combine the Fisher Sensitivity \(S_l\) and Effective Rank \(R_l\) into a single importance score \(Q_l\):
% \begin{equation}
% Q_l = (S_l)^\beta \times (R_l)^{1-\beta},
% \end{equation}
% where \(\beta\) is a hyperparameter that controls the relative importance of sensitivity and effective rank. 

% Given normalized importance scores \(Q_l\) for each layer and target global compression ratio \(k\%\), the compression ratios \(k_l\) are computed as:
% $
% k_l = k \cdot \frac{1 - Q_l}{\sum_{j=1}^L 1 - Q_j}
% $.
% As shown in Fig.\ref{heuristic}, the derived compression ratios $Q_l$ effectively capture both the sensitivity trends ($S_l$) and compressibility patterns ($R_l$) across layers, clearly revealing each layer's relative importance. 

\textbf{Combining Sensitivity and Effective Rank.} 
Layers with higher sensitivity values are more critical to the model’s performance and should be preserved more aggressively. Similarly, layers with lower effective ranks are more compressible and can tolerate higher compression. To assign compression ratios to each layer, we first combine the Fisher Sensitivity \( S_l \) and Effective Rank \( R_l \) into a unified importance score \( Q_l \):
\begin{equation}
Q_l = (S_l)^\beta \cdot (R_l)^{1 - \beta}, \tag{12}
\end{equation}
where \( \beta \) is a hyperparameter that controls the relative importance of sensitivity and effective rank.

Given normalized importance scores \( Q_l \) for each layer and a target global compression ratio \( k \), we define the per-layer preservation ratios \( p_l \) (i.e., the proportion of parameters retained in each layer) such that the average preservation ratio across all layers equals \( 1 - k \). The preservation ratios are computed as:

\begin{equation}
p_l = \frac{Q_l}{\sum_{j=1}^{L} Q_j} \cdot L \cdot (1 - k). \tag{13}
\end{equation}

\begin{figure}[htbp]
\centering
\includegraphics[width=1.0\columnwidth]{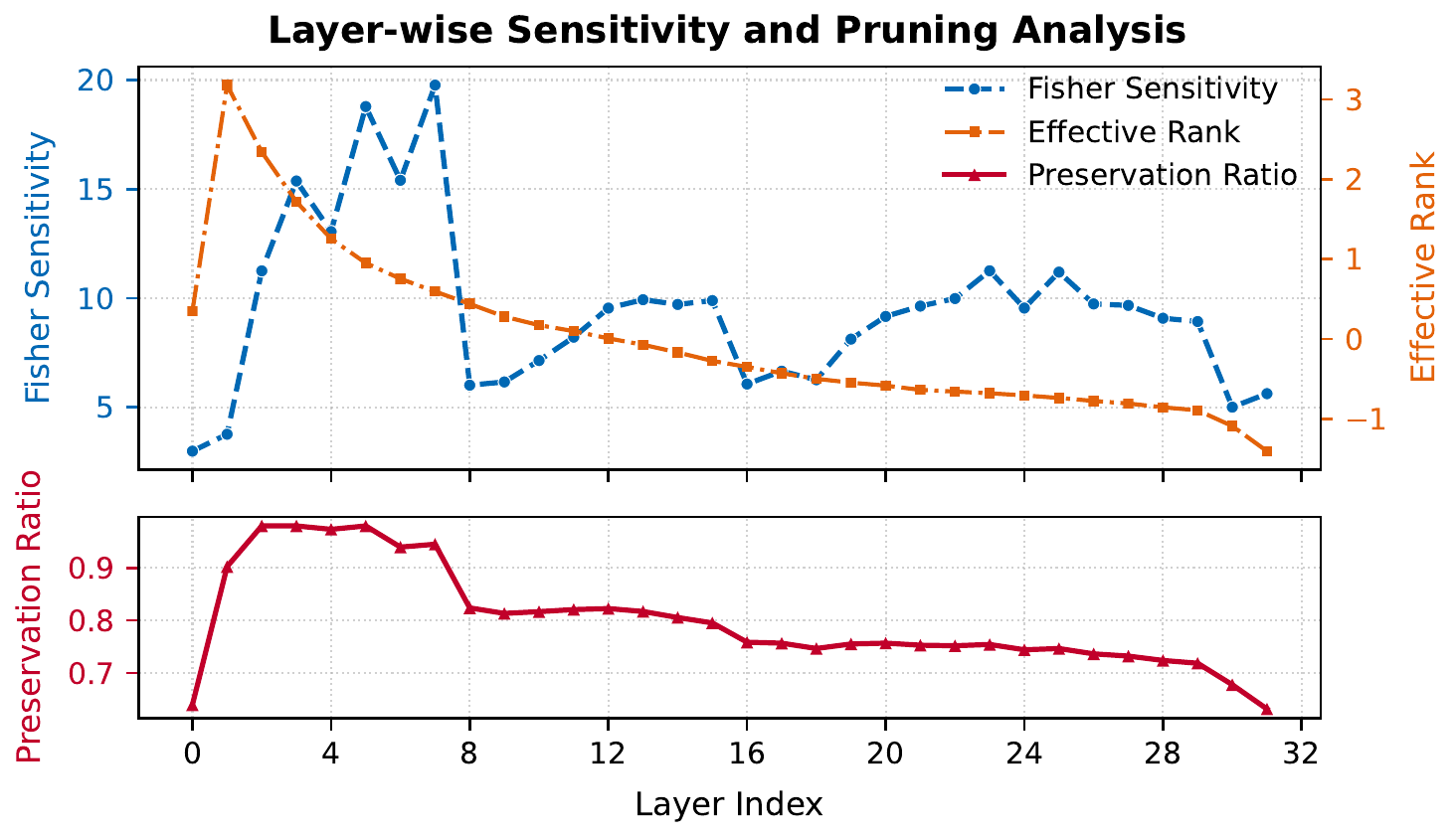}
\caption{The $Q_l$, $S_l$, and $R_l$ values for each layer in Vicuna-7B at 20\% compression ratio.}
\label{heuristic}
\end{figure}

The corresponding compression ratios are then given by \( 1 - p_l \). This formulation ensures that layers with higher importance scores \( Q_l \) are compressed less aggressively, while maintaining the desired global compression budget. As shown in Fig.\ref{heuristic}, by capturing both the sensitivity trends \( S_l \) and compressibility patterns \( R_l \), the derived preservation ratios clearly reflect each layer’s relative contribution to model performance.

\begin{table*}[h]
\centering
\renewcommand{\arraystretch}{1.3}
\resizebox{1.0\textwidth}{!}{%
% \begin{tabular}{cc|cccc|cccc|cccc|cccc}
\begin{tabular}{cc|ccc|cccccccc}
\hline \hline
Model & Method & WikiText-2 & PTB & C4 & Openb. & ARC\_e & WinoG. & HellaS. & ARC\_c & PIQA & MathQA & Average \\ \hline
\multirow{4}{*}{LLaMA-7B} & ASVD & 95.268 & 200.937 & 86.269 & 0.186 & 0.379 & 0.557 & 0.333 & 0.242 & 0.607 & 0.218 & 0.360 \\
& FWSVD & 33.001 & 53.587 & 38.240 & 0.186 & 0.507 & 0.572 & 0.343 & 0.242 & 0.632 & 0.217 & 0.386 \\
& SVD-LLM & 9.526 & 28.967 & 26.390 & 0.242 & 0.509 & 0.570 & 0.352 & 0.269 & 0.630 & 0.227 & 0.400 \\
& Ours & \textbf{9.427} & \textbf{22.270} & \textbf{19.909} & \textbf{0.242} & \textbf{0.602} & \textbf{0.640} & \textbf{0.405} & \textbf{0.296} & \textbf{0.661} & \textbf{0.230} & \textbf{0.440} \\ \hline
\multirow{4}{*}{Vicuna-7B} & ASVD & 91.388 & 415.615 & 136.157 & 0.158 & 0.335 & 0.503 & 0.287 & 0.208 & 0.556 & 0.205 & 0.322 \\
& FWSVD & 43.690 & 239.318 & 64.753 & 0.172 & 0.459 & 0.545 & 0.312 & 0.224 & 0.613 & 0.221 & 0.364 \\
& SVD-LLM & 12.416 & 124.506 & 39.528 & 0.244 & 0.506 & 0.570 & 0.353 & 0.270 & 0.629 & 0.228 & 0.400 \\
& Ours & \textbf{12.144} & \textbf{81.089} & \textbf{28.837} & \textbf{0.248} & \textbf{0.573} & \textbf{0.597} & \textbf{0.384} & \textbf{0.293} & \textbf{0.659} & \textbf{0.232} & \textbf{0.427} \\ \hline
\multirow{4}{*}{DeepSeek-7B} & ASVD & 85.169 & 87.709 & 79.853 & 0.154 & 0.390 & 0.516 & 0.312 & 0.213 & 0.610 & 0.210 & 0.344 \\
& FWSVD & 68.416 & 99.775 & 118.319 & 0.142 & 0.406 & 0.551 & 0.296 & 0.194 & 0.595 & 0.220 & 0.344 \\
& SVD-LLM & 10.841 & 30.747 & 32.622 & 0.260 & 0.589 & 0.609 & 0.384 & 0.283 & 0.670 & 0.232 & 0.432 \\
& Ours & \textbf{9.895} & \textbf{20.977} & \textbf{22.558} & \textbf{0.276} & \textbf{0.628} & \textbf{0.631} & \textbf{0.415} & \textbf{0.312} & \textbf{0.700} & \textbf{0.239} & \textbf{0.457} \\
\hline \hline
\end{tabular}
}
\caption{Performance of LLaMA-7B, Vicuna-7B and Deepseek-7B models compressed by DipSVD and baselines at 30\% compression ratio, evaluated on three language modeling datasets (measured by perplexity) and seven classification datasets (measured by average accuracy). The best performance for each case is marked in bold.}
\label{performance1}
\end{table*}

\subsection{Integrated Compression Process}
DipSVD incorporates local and global significance protection mechanisms into compression:

\textbf{Step 1: Layer Whitening.} For each layer, perform data whitening using the matrix $\tilde{X}^T \tilde{X}$ and SVD, as described in Sec.3.1.1.

\textbf{Step 2: Layer-Specific Compression Ratios.} Use Bayesian optimization (or the heuristic-based approach) to determine the optimal compression ratios \(k_1, k_2, \ldots, k_L\) for each layer, as described in Sec.3.2.

\textbf{Step 3: SVD Low-Rank Approximation.} For each layer \(l\), apply the compression ratio \(k_l\) to the weight matrix \(W_l\) using SVD-based low-rank approximation:

    1) Perform SVD on the whitened weight matrix \(W_l S\):
    \begin{equation}
       W_l S = U_l \Sigma_l V_l^{T},
    \end{equation}
    where \(U_l \in \mathbb{R}^{n \times r}\) and \(V_l \in \mathbb{R}^{m \times r}\) are orthogonal matrices, and \(\Sigma_l \in \mathbb{R}^{r \times r}\) is a diagonal matrix containing the singular values \(\sigma_1, \sigma_2, \ldots, \sigma_r\) of \(W_l S\).
    
    2) Truncate the smallest singular values in \(\Sigma_l\) based on \(k_l\), obtaining the truncated diagonal matrix \(\text{Trunc}_*(\Sigma_l)\).
    
    3) Construct the compressed weight matrix \(W_l'\):
\begin{equation}
   W_l' = U_l \times \text{Trunc}_*(\Sigma_l) \times V_l^{T} \times S^{-1}.
\end{equation}

     4) To further reduce memory usage, replace the original weight matrix \(W_l\) with two low-rank matrices \(W_{u,l} \in \mathbb{R}^{n \times \tilde{r}}\) and \(W_{v,l} \in \mathbb{R}^{\tilde{r} \times m}\):
\begin{equation}
   W_{u,l} = U_l \times [\text{Trunc}_*(\Sigma_l)]^{1/2},
\end{equation}
\begin{equation}
   W_{v,l} = [\text{Trunc}_*(\Sigma_l)]^{1/2} \times V_l^{T} \times S^{-1},
\end{equation}
   where \(\tilde{r}\) is the rank after truncation.

\section{Experiments}
\subsection{Experimental setup}
\textbf{Foundation LLMs.} We conducted experiments on existing popular LLMs at various scales, including LLaMA-\{7B, 13B\} \citep{touvron2023llama}, Vicuna-\{7B, 13B\}-v1.5 \citep{chiang2023vicuna} and Deepseek-{7B} \citep{deepseekai2025deepseekv3technicalreport}.

\textbf{Baselines.} We compared DipSVD with several previous SVD-based compression methods including FWSVD \citep{hsu2022language}, ASVD \citep{yuan2023asvd} and SVD-LLM \citep{wang2024svd}.

\textbf{Benchmarks.} We measure zero-shot accuracy on commonsense reasoning datasets (i.e., PIQA \citep{bisk2020piqa}, HellaSwag \citep{zellers2019hellaswag}, WinoGrande \citep{sakaguchi2021winogrande}, ARC-easy \citep{clark2018think}, ARC-challenge \citep{clark2018think}, MathQA\citep{amini2019mathqainterpretablemathword} and OpenbookQA \citep{mihaylov2018can}) using the lm-evaluation-harness package \citep{eval-harness}.  To assess sequence prediction performance, we report perplexity for DipSVD and the baselines on WikiText-2 \citep{merity2016pointersentinelmixturemodels}, PTB \citep{marcus-etal-1993-building} and C4 \citep{mihaylov2018suitarmorconductelectricity}. 

\textbf{Implementation Details.} To ensure a fair comparison, we followed ASVD \citep{yuan2023asvd} to randomly select 256 samples from WikiText-2 as the calibration data. All of our experiments are conducted on NVIDIA A100 GPUs.

\subsection{Overall Performance}

We evaluate the overall performance of DipSVD from three aspects: (1) performance on different LLMs, (2) performance on LLMs with larger scales, (3) performance under different compression ratios. Detailed results, contents generated by the compressed LLMs, and an analysis of the computational gains achieved by our method are provided in the Appendix.

\textbf{Performance on different LLMs.} We first compare the zero-shot task performance and perplexity metrics between DipSVD and the baseline on three different LLMs, including LLaMA-7B, Vicuna-7B and DeepSeek-7B under 30\% compression ratio. As shown in Tab.\ref{performance1}, DipSVD consistently outperforms the baselines across all three LLMs and all ten datasets. Specifically, we achieve the lowest perplexity (9.427) on LLaMA-7B model. For downstream tasks, our method attains the highest average accuracy (0.457 vs. SVD-LLM’s 0.432 on DeepSeek-7B), with notable gains on reasoning-heavy benchmarks like ARC-Challenge (+4.3\%). These results collectively show that DipSVD effectively reduces model size and complexity while better preserving model performance compared to existing methods. The consistent improvements across different model architectures further demonstrate the robustness and generalizability of our proposed method. 

\begin{table}[ht]
\centering
\renewcommand{\arraystretch}{1.3}
\resizebox{1.0\columnwidth}{!}{%
\begin{tabular}{cc|cccc}
\hline \hline
Model & Method & WikiText-2 & PTB & C4 & Average \\
\hline
\multirow{4}{*}{LLaMA-13B} & ASVD & 17.648 & 32.963 & 20.866 & 0.425 \\
& FWSVD & 12.963 & 22.123 & 18.509 & 0.383 \\
& SVD-LLM & \textbf{7.618} & 17.823 & 18.825 & 0.449 \\
& Ours & 7.697 & \textbf{15.681} & \textbf{14.614} & \textbf{0.472} \\
\hline
\multirow{4}{*}{Vicuna-13B} & ASVD & 28.309 & 637.196 & 39.799 & 0.401 \\
& FWSVD & 32.715 & 310.304 & 47.408 & 0.383 \\
& SVD-LLM & 9.616 & 145.715 & 29.204 & 0.449 \\
& Ours & \textbf{9.070} & \textbf{49.948} & \textbf{19.203} & \textbf{0.458} \\
\hline \hline
\end{tabular}
}
\caption{Perplexity on three language modeling datasets and average accuracy of seven datasets of LLaMA-13B and Vicuna-13B at 30\% compression ratio.}
\label{performance2}
\end{table}

\begin{figure}[h]
\centering
\includegraphics[width=1.0\columnwidth]{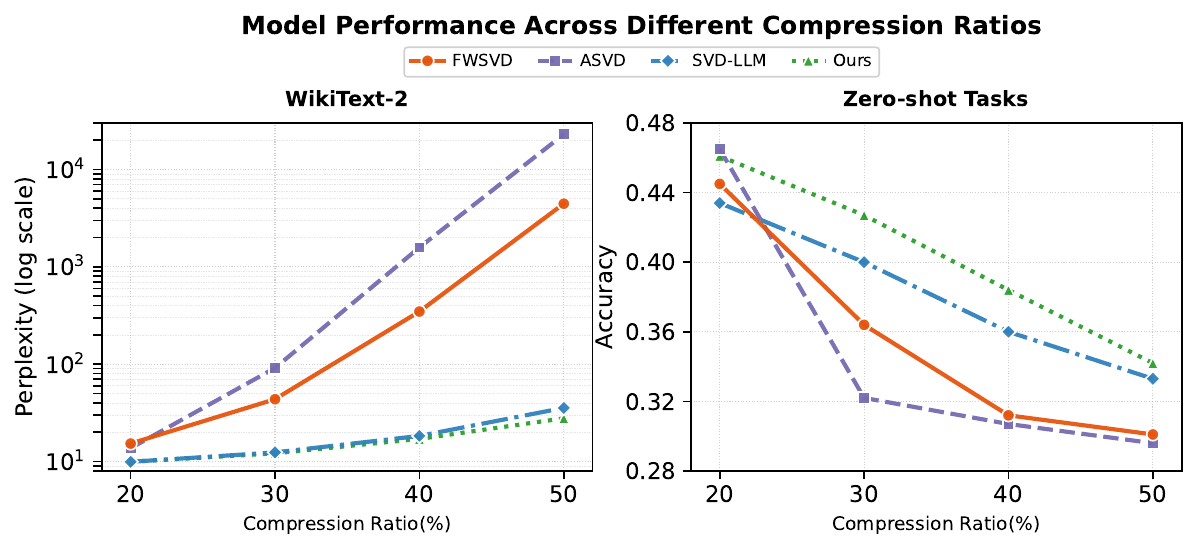}
\caption{Perplexity on WikiText-2 and average zero-shot accuracy of Vicuna-7B compressed by DipSVD and baselines under 20\% to 50\% compression ratios.}
\label{performance3}
\end{figure}

\textbf{Performance on LLMs with larger scales.} We compare the performance between DipSVD and the baselines on LLaMA-13B and Vicuna-13B under 30\% compression ratio on ten datasets in Tab.\ref{performance2}. Specifically, our DipSVD method reduce PTB perplexity by 65.7\% compared to SVD-LLM (49.948 vs 145.715) for Vicuna-13B. The results demonstrate that our method effectively scales to LLMs while maintaining superior capability preservation compared to existing compression methods.

\textbf{Performance under Different Compression Ratios.} We compare the performance between DipSVD and the baselines on Vicuna-7B under compression ratio ranging from 20\% to 50\% on WikiText-2 datasets and seven classification datasets. As shown in Fig.\ref{performance3}, DipSVD consistently outperforms all baselines, and the performance gap gain with the compression ratio increases.

\subsection{Performance of heuristics method}

The heuristic method also serves as a key strategy for global importance preservation. Tab.\ref{heuristics} compares perplexity scores across SVD-LLM, Bayesian optimization method, and the heuristic method. The results demonstrate that the heuristic method (DipSVD\_H) significantly outperforms SVD-LLM across multiple compression ratios on all three datasets. While the Beyesian optimization method (DipSVD\_B) achieves the best performance with large computational overheads, DipSVD\_H remains competitive while offering a practical trade-off between performance and efficiency.

\begin{table}[h]
\centering
\renewcommand{\arraystretch}{1.2}
\resizebox{1.0\columnwidth}{!}{%
\begin{tabular}{ccccc}
\hline \hline
Compression Ratio& Method& WikiText2& PTB& C4\\ \hline
\multirow{3}{*}{0.2} & SVD-LLM & \textbf{9.942}& 71.366& 23.358 \\
 & DipSVD\_B& 9.952& 56.869&\textbf{19.722}\\
 & DipSVD\_H& 9.988& \textbf{54.380}& 19.950\\ \hline
\multirow{3}{*}{0.3} & SVD-LLM & 12.416& 124.506 & 39.528\\
 & DipSVD\_B& \textbf{12.144}& \textbf{81.089}&\textbf{28.837}\\
 & DipSVD\_H& 12.378& 82.872& 30.748\\ \hline
\multirow{3}{*}{0.4} & SVD-LLM & 18.346& 261.100& 77.706\\
 & DipSVD\_B& \textbf{17.085}& \textbf{142.752}&\textbf{49.183}\\
 & DipSVD\_H& 17.290& 168.404& 54.794\\ \hline
\multirow{3}{*}{0.5} & SVD-LLM & 35.569& 615.591& 185.780\\
 & DipSVD\_B& \textbf{27.807}& \textbf{375.093}&\textbf{111.996}\\
 & DipSVD\_H& 30.180& 390.512& 118.873\\ \hline \hline
\end{tabular}
}
\caption{Heuristics result of compressed Vicuna-7B. DipSVD\_B represents compression model by Bayesian optimization method. DipSVD\_H represents compression model by the heuristic method.}
\label{heuristics}
\end{table}

\textbf{Connection with Bayesian optimization.}
In the experiments, the heuristic method showed the strongest correlation with the optimization-based method when $\beta = 0.25$. Therefore, we uniformly set $\beta$ to 0.25. Tab.\ref{pearson} shows the Pearson correlation coefficient between the layer-wise compression rates obtained by heuristic and Bayesian optimization across different target compression ratios. The correlation coefficients consistently exceed 0.64, indicating strong agreement between the two methods at all compression levels.

\begin{table}[h]
\centering
\resizebox{1.0\columnwidth}{!}{%
\renewcommand{\arraystretch}{1.3}
\begin{tabular}{ccccc}
\hline \hline
\multicolumn{1}{c|}{Compression Ratio} & 0.2 & 0.3 & 0.4 & 0.5 \\ \hline
\multicolumn{1}{c|}{Pearson Correlation Coefficient} & 0.645 & 0.669 & 0.711 & 0.706 \\ \hline \hline
\end{tabular}
}
\caption{Pearson correlation coefficients between heuristic and Bayesian optimization methods across different compression ratios.}
\label{pearson}
\end{table}

\subsection{Ablation Study}

\subsubsection{Hyperparametric Ablation Experiment}

\begin{figure}[ht]
\centering
\includegraphics[width=1.0\columnwidth]{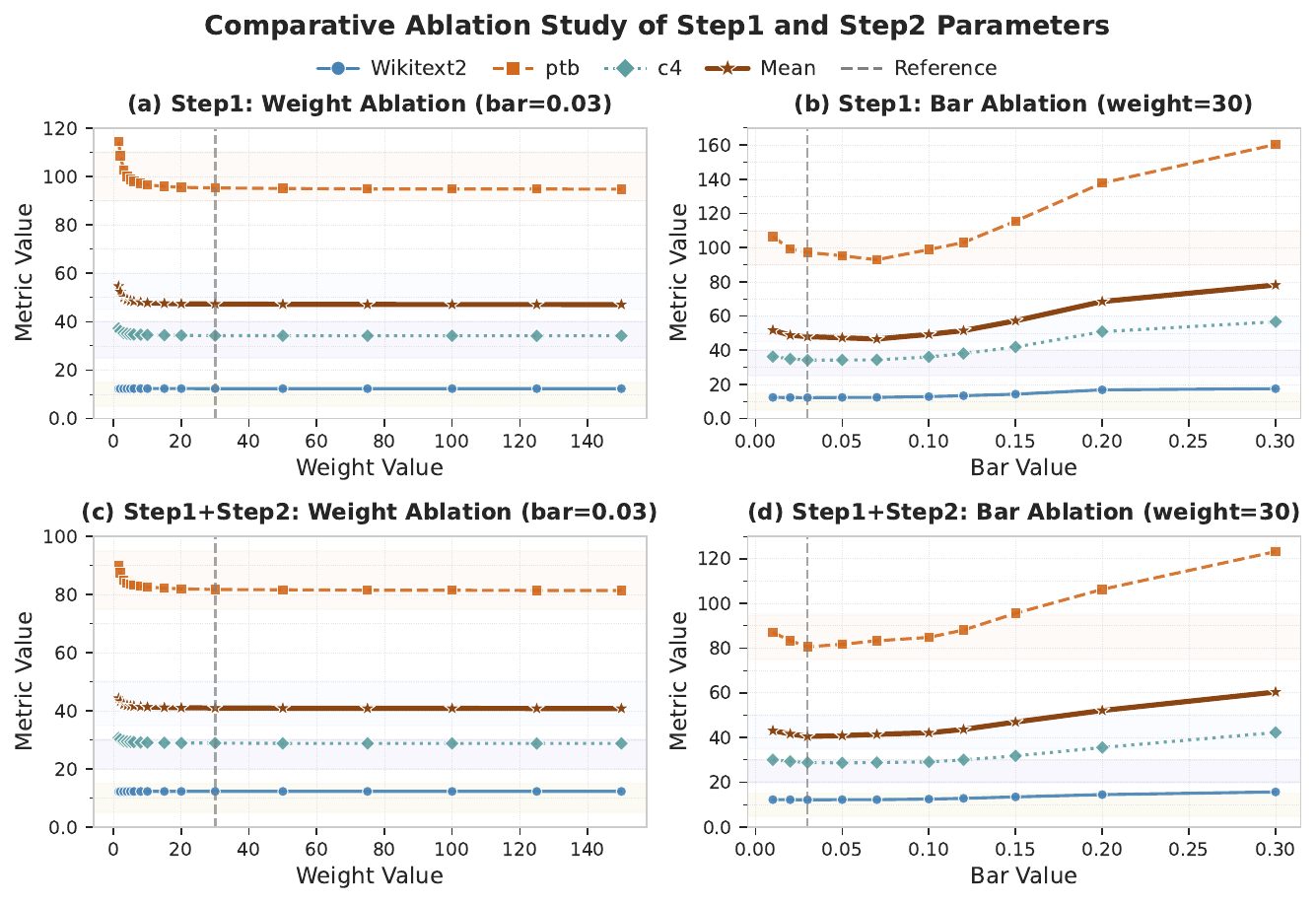}
\caption{Parameter ablation studies: (a) Step1 weight ablation (bar=0.03), (b) Step1 bar ablation (weight=30), (c) Full DipSVD weight ablation (bar=0.03), (d) Full DipSVD bar ablation (weight=30). Dashed lines mark reference valus (weight=30, bar=0.03).}
\label{ablation}
\end{figure}

We conducted ablation experiments to evaluate the impact of two hyperparameters in DipSVD:  weight (importance amplification factor) and bar (top channel selection ratio). The results are shown in Fig.\ref{ablation}. First, we isolated the effect of weight by testing progressive values from 0 to 150 while fixing bar=0.03, revealing that higher weights introduce better performance as the parameters increase until it remains unchanged. Next, we analyzed bar with values from 0 to 0.3, demonstrating that the performance will be better and then worse as the parameters increase. The optimal performance of both DipSVD (Step1) and DipSVD (Step1+Step2) can be achieved when weight=30 and bar=0.03.

\subsubsection{Modular Sensitivity Study}
We conduct ablation studies to evaluate our DipSVD method by: (1) Isolating Step1 (local importance protection with uniform layer compression ratios), (2) Isolating Step2 (global importance protection with uniform SVD compression) and (3) Combining Step1 and Step2 (full DipSVD). This comparison quantifies the impact of local importance protection (Step1) and how global importance protection (Step2) enhances the base performance achieved by local compression (Step1).

\begin{table}[h]
\centering
\resizebox{1.0\columnwidth}{!}{%
\begin{tabular}{c|c|ccc}
\hline \hline
Method& Hyper& WikiText2& PTB& C4\\ \hline
Vanilla & None & 6.7836 & 30.853 & 9.2064 \\
SVD-LLM & None & 12.4212 & 124.6766 & 39.5712 \\
DipSVD (Step1) & weight 30 bar 0.03 & 12.2663 & 97.2792 & 34.2737 \\
DipSVD (Step2) & None & 12.1868 & 94.9336 & 31.7530 \\
DipSVD (Step1+Step2) & weight 30 bar 0.03 & 12.1578 & 80.6209 & 28.8224 \\ 
\hline \hline
\end{tabular}
}
\caption{Perplexity of compressed Vicuna-7B: DipSVD (Step1) preserves local importance only, DipSVD (Step2) preserves global importance only, while DipSVD (Step1+Step2) preserves both aspects.}
\label{step ablation}
\end{table}

\begin{figure}[h]
\centering
\includegraphics[width=1.0\columnwidth]{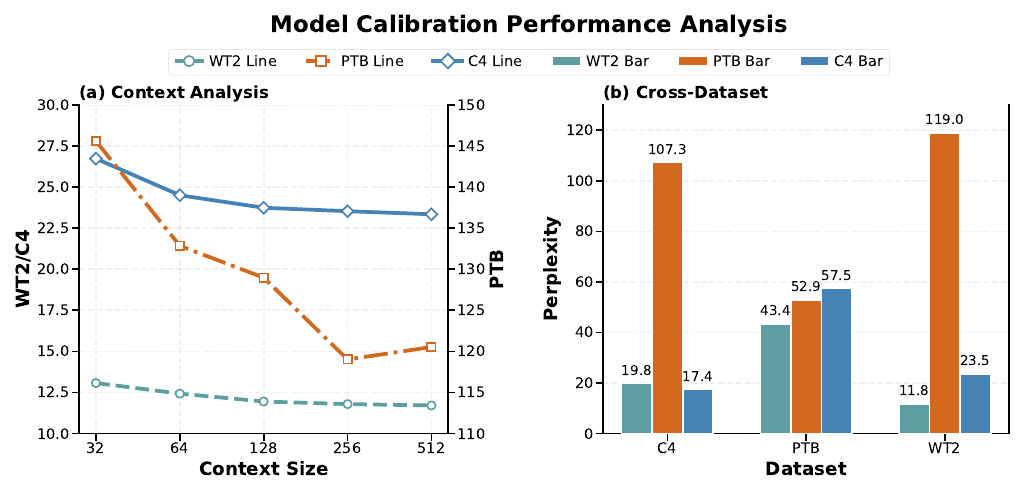}
\caption{Ablation study of calibration datasets: (a) Perplexity with varying sizes of Wikitext2 calibration data, (b) Perplexity with different calibration datasets.}
\label{calibration}
\end{figure}

From Tab.\ref{step ablation}, we observe that: (1) Both DipSVD (Step1) and DipSVD (Step2) outperforms SVD-LLM on all datasets, confirming the effectiveness of our proposed local and global importance protection; (2) The full DipSVD (Step1+Step2) further reduces perplexity substantially, demonstrating the synergistic effect of combining both protection strategies in SVD compression.

\subsubsection{Impact of Calibration Data.}
We examine the impact of calibration data used for channel-weighted data whitening. Fig.\ref{calibration} shows the performance of compressed Vicuna-7B when changing the choice and size of the calibration data. The results show that increased context size will improve the final performance of the compression model to a certain extent, and will work best when the test set is consistent with the calibration set.

\section{Conclusion}
In this paper, we propose DipSVD, a compression framework that jointly considers both local and global importance factors in large languge models (LLMs) to achieve more efficient and balanced compression. Specifically, we introduce a channel-weighted data whitening technique to preserve the most critical singular vectors within each weight matrix and develop two strategies for determining layer-specific compression ratios, enabling less important layers to absorb a greater portion of the compression burden. Extensive experiments across 10 datasets, 5 models from 3 LLM families, and 4 compression scales consistently demonstrate that DipSVD achieves superior performance compared to existing SVD-based compression methods.

\section*{Limitations}

While DipSVD demonstrates strong performance across various models, datasets, and compression scales, several limitations remain. 

First, the channel-weighted data whitening procedure employs a fixed amplification factor across all layers and channels, which may not optimally adapt to the varing statistical properties of different layers or channel distributions. A more dynamic or data-driven adjustment of the weighting factors could potentially further enhance compression performance.

Second, the assignment of layer-wise compression ratios in DipSVD relies on handcrafted importance metrics based on Fisher sensitivity and effective rank. While this design is computationally efficient and effective in practice, it may not fully capture complex inter-layer dependencies or feature interactions. Developing more sophisticated global importance modeling techniques could potentially lead to further improvements.

Nevertheless, despite these limitations, DipSVD consistently outperforms existing SVD-based compression methods across diverse evaluation settings, demonstrating its robustness and practical effectiveness.

% \section*{Acknowledgments}

% This document has been adapted
% by Steven Bethard, Ryan Cotterell and Rui Yan
% from the instructions for earlier ACL and NAACL proceedings, including those for
% ACL 2019 by Douwe Kiela and Ivan Vuli\'{c},
% NAACL 2019 by Stephanie Lukin and Alla Roskovskaya,
% ACL 2018 by Shay Cohen, Kevin Gimpel, and Wei Lu,
% NAACL 2018 by Margaret Mitchell and Stephanie Lukin,
% Bib\TeX{} suggestions for (NA)ACL 2017/2018 from Jason Eisner,
% ACL 2017 by Dan Gildea and Min-Yen Kan,
% NAACL 2017 by Margaret Mitchell,
% ACL 2012 by Maggie Li and Michael White,
% ACL 2010 by Jing-Shin Chang and Philipp Koehn,
% ACL 2008 by Johanna D. Moore, Simone Teufel, James Allan, and Sadaoki Furui,
% ACL 2005 by Hwee Tou Ng and Kemal Oflazer,
% ACL 2002 by Eugene Charniak and Dekang Lin,
% and earlier ACL and EACL formats written by several people, including
% John Chen, Henry S. Thompson and Donald Walker.
% Additional elements were taken from the formatting instructions of the \emph{International Joint Conference on Artificial Intelligence} and the \emph{Conference on Computer Vision and Pattern Recognition}.

% Bibliography entries for the entire Anthology, followed by custom entries
%\bibliography{anthology,custom}
% Custom bibliography entries only
\bibliography{custom}

\appendix

\section{Pseudocode of DipSVD}

Algorithm \ref{algorithm1} shows the pseudocode of DipSVD. Before compression, DipSVD randomly collects a small amount of sentences as the calibration data C, it then runs the truncation-aware data whitening process as shown in Algorithm \ref{algorithm2} to obtain the set of whitening matrix  $\text{Set}_S$ for the weight to compress. After that, it runs the SVD and truncation with $\text{Set}_S$ on each weight matrix in the LLM. Before formally compressing the model, it is necessary to use heuristic methods or Bayesian optimization methods to obtain the compression ratios of different layers, as shown in Algorithm \ref{algorithm3}.

\begin{algorithm*}[ht]
    \caption{Pseudocode of DipSVD}
    \textbf{Input}: Original model $M$, Target compression ratio $R$ \\
    \textbf{Output}: Compressed model $M'$ \\
    \textbf{Procedure}: DipSVD($M$, $R$)
    
    \begin{algorithmic}[1] %[1] enables line numbers
        \State Randomly collect several sentences as the calibration data $C$
        \State $S \gets \text{CHANNEL-WEIGHTED DATA WHITENING}(M,C)$ \Comment{Extract the whitening matrix} 
        \State $\{k_1, k_2, ...,k_L\} \gets \text{LAYER-SPECIFIC COMPRESSION}(M)$ \hfill{$\triangleright$ Extract the layer specific compression ratios}
        \For{each layer $l \in \{1,...,L\}$}
            \State $W_l \gets \text{GetLayerWeights}(\mathcal{M}, l)$ \hfill \text{$\triangleright$ Obtain the set of weights in $M$ to compress}
            \State $U_l,\Sigma_l, V_l \gets \text{SVD}(W_lS_l)$ \hfill {$\triangleright$ Apply singular value decomposition on W}
            \State $k_l \gets \arg\min_k \left(\frac{\sum_{i=1}^k \sigma_i^2}{\sum \sigma_i^2} \geq k_l\right)$
            \State $\Sigma_l' \gets \text{diag}(\sigma_1,...,\sigma_{k_l},0,...,0)$  \hfill {$\triangleright$ Truncate the smallest singular values in $\Sigma$}
            \State $W_l' \gets U_l \Sigma_l' V_l S_l^{-1}$
            \State Replace $W_l$ with $ W_l'$ in $M'$
        \EndFor
        \State \textbf{return} $M'$ 
    \end{algorithmic}
\textbf{End Procedure}
\label{algorithm1}
\end{algorithm*}

\begin{algorithm*}[ht]
    \caption{Pseudocode of Channel-weighted data whitening}
    \textbf{Input}: Original model $M$; Calibration Data $C$; Diagonal scaling matrix $D$ \\
    \textbf{Output}: Set of whitening matrices in $M$ for the weight to compress  $\text{Set}_S$\\
    \textbf{Procedure}: CHANNEL-WEIGHTED DATA WHITENING($M$,$C$)
    
    \begin{algorithmic}[1] %[1] enables line numbers
        \State $\text{Set}_S \gets \phi$ \hfill \text{$\triangleright$ Initialize the set of whitening matrices}
        \State $\text{Set}_{W} \gets M$ \hfill \text{$\triangleright$ Obtain the set of weights in $M$ to compress}
        \For{$W \in \text{Set}_S$}
            \State $X \gets M(W,C)$\; \hfill{$\triangleright$ Obtain the input activation of the weight matrix $W$}
            \State $\tilde{X}^T \tilde{X} \gets D^TX^TXD$ \hfill{$\triangleright$ Obtain the input activation of the weight matrix $W$}
            \State $S \gets \text{SVD}(\tilde{X}^T\tilde{X})$ \hfill {$\triangleright$ Apply singular value decomposition on $\tilde{X}^T\tilde{X}$}
            \State $\Sigma_1 \gets \text{Trunc.}(\Sigma)$ \hfill {$\triangleright$ Truncate the smallest singular values in $\Sigma$}
            \State $\text{Set}_{S} \gets S \bigcup \text{Set}_{S}$ \hfill {$\triangleright$ Store the whitening weight matrix in the set}
        \EndFor
        \State \textbf{return} $\text{Set}_S$
    \end{algorithmic}
\textbf{End Procedure}
\label{algorithm2}
\end{algorithm*}

\begin{algorithm*}[ht]
    \caption{Pseudocode of Layer-specific Compression}
    \textbf{Input}: Original model $M$; Input activation $x$; Target compression ratio $R$; \

    \textbf{Parameters}: Global importance preservation method $m \in \{\text{Bayesian}, \text{Heuristic}\}$, Energy threshold $\tau \in (0,1)$ (default: 0.95), Trade-off parameter $\beta \in [0,1]$ (default: 0.3)\\
    \textbf{Output}: A list of allocated compression ratios  $\{k_1,...,k_L\}$  \begin{algorithmic}[1] %[1] enables line numbers
        \State Randomly collect several sentences as the calibration data $C$
        \If{$m = \text{Heuristic}$}
            \State $S \gets \text{FisherSensitivity}(M, C)$ \Comment{Algorithm~\ref{algorithm4}}
            \State $U \gets \text{EffectiveRank}(M, C, \tau)$ \Comment{Algorithm~\ref{algorithm5}}
    
            \State Normalize metrics:
     $\tilde{S} \gets \frac{S-\min(S)}{\max(S)-\min(S)} + \epsilon$
    $\tilde{U} \gets \frac{U-\min(U)}{\max(U)-\min(U)} + \epsilon$
    
            \State Compute combined importance:
    $W \gets \tilde{S}^\beta \circ \tilde{U}^{1-\beta}$
    
             \State Allocate ratios:
    $\{k_1,...,k_L\} \gets \text{ProportionalAllocation}(W, R)$
        \Else
        \State Initialize Bayesian Optimizer:
    $\mathcal{B} \gets \text{BO}(\text{domain}=[0.25,1]^L, \text{acq}=\text{EI})$
    
        \For{$t \gets 1$ to $T$}
            \State $\mathbf{r}_t \gets \mathcal{B}.\text{query}()$
            \State $M' \gets \text{Compress}(M, \mathbf{r}_t)$ 
            \State $\text{score} \gets -\text{Perplexity}(M', C)$
            \State $\mathcal{B}.\text{update}(\mathbf{r}_t, \text{score})$
        \EndFor
    
        \State $\{k_1,...,k_L\} \gets \mathcal{B}.\text{best\_params}()$
    \EndIf
    \State \textbf{return} $\{k_1,...,k_L\}$
    \end{algorithmic}
\label{algorithm3}
\end{algorithm*}

\begin{algorithm*}[ht]
\caption{Fisher Sensitivity Computation}
\begin{algorithmic}[1]
\Function{ComputeFisherSensitivity}{$M, C$}
    \State Initialize $S \gets \mathbf{0}_L$
    \For{each batch $B \subset C$}
        \State Compute gradients $\nabla\mathcal{L}$ via backpropagation
        \For{each layer $l \in \{1,...,L\}$}
            \State $s_l \gets \sum_{p\in\theta_l}\frac{\|\nabla_p\mathcal{L}\|_2}{\|p\|_2 + \epsilon}$
            \State $S[l] \gets S[l] + s_l$
        \EndFor
    \EndFor
    \State Apply segmented normalization to $S$
    \State \Return $S^{-1}$ \Comment{Inverse sensitivity}
\EndFunction
\end{algorithmic}
\label{algorithm4}
\end{algorithm*}

\begin{algorithm*}[ht]
\caption{Effective Rank Computation}
\begin{algorithmic}[1]
\Function{ComputeEffectiveRank}{$M, \mathcal{C}, \tau$}
    \State Initialize $U \gets \mathbf{0}_L$
    \For{each batch $B \subset C$}
        \State Get hidden states $\{h_1,...,h_L\}$
        \For{each layer $l \in \{1,...,L\}$}
            \State $H_l \gets \text{reshape}(h_l, [-1, d_l])$
            \State $\{\sigma_i\} \gets \text{SVD}(H_l)$
            \State $k \gets \min\{k' | \sum_{i=1}^{k'}\sigma_i^2 \geq \tau\sum\sigma_i^2\}$
            \State $U[l] \gets U[l] + k$
        \EndFor
    \EndFor
    \State $U \gets \frac{U}{|\mathcal{C}|/\text{batch\_size}}$
    \State \Return $\text{z-score}(U)$
\EndFunction
\end{algorithmic}
\label{algorithm5}
\end{algorithm*}

\section{Analysis of model computational complexity}
Computational complexity mainly depends on the model structure, parameter quantity, sequence length and hardware implementation. This section takes Vicuna-7b model as an example to discuss the computational gain brought by the DipSVD method at a 40\% compression ratio.

The DipSVD method achieves significant computational savings in Vicuna-7B through structured low-rank approximation of weight matrices. For a given weight matrix $W \in \mathbb{R}^{m\times n}$, we decompose it via truncated SVD as $W \approx U_l \Sigma_k V^T_k$, where $k \ll \min(m,n)$ is the target rank detemined by Bayesian optimization under a 60\% parameter budget constraint ($mk +kn = 0.6mn$). This decomposition transforms the original matrix multiplication $WX$ (computational complexity $C_{WX} = m\cdot n\cdot p$ ) into a two-step operation: (1) $\hat{X} = \Sigma_k V_k^T X $ ($knp$ FLOPs) followed by (2) $U_k \hat{X}$ ($mkp$ FLOPs), yielding total complexity $C_{UV} = k \cdot p(m+n)$. Substituing the optimal rank $k= \frac{0.6mn}{m+n}$ derived from the parameter constraint, we obtain a theoetical FLOPs reduction ratio of $r= 1-\frac{C_{UV}}{C_{WX}}=40\%$.

\begin{figure}[h]
\centering
\includegraphics[width=0.9\columnwidth]{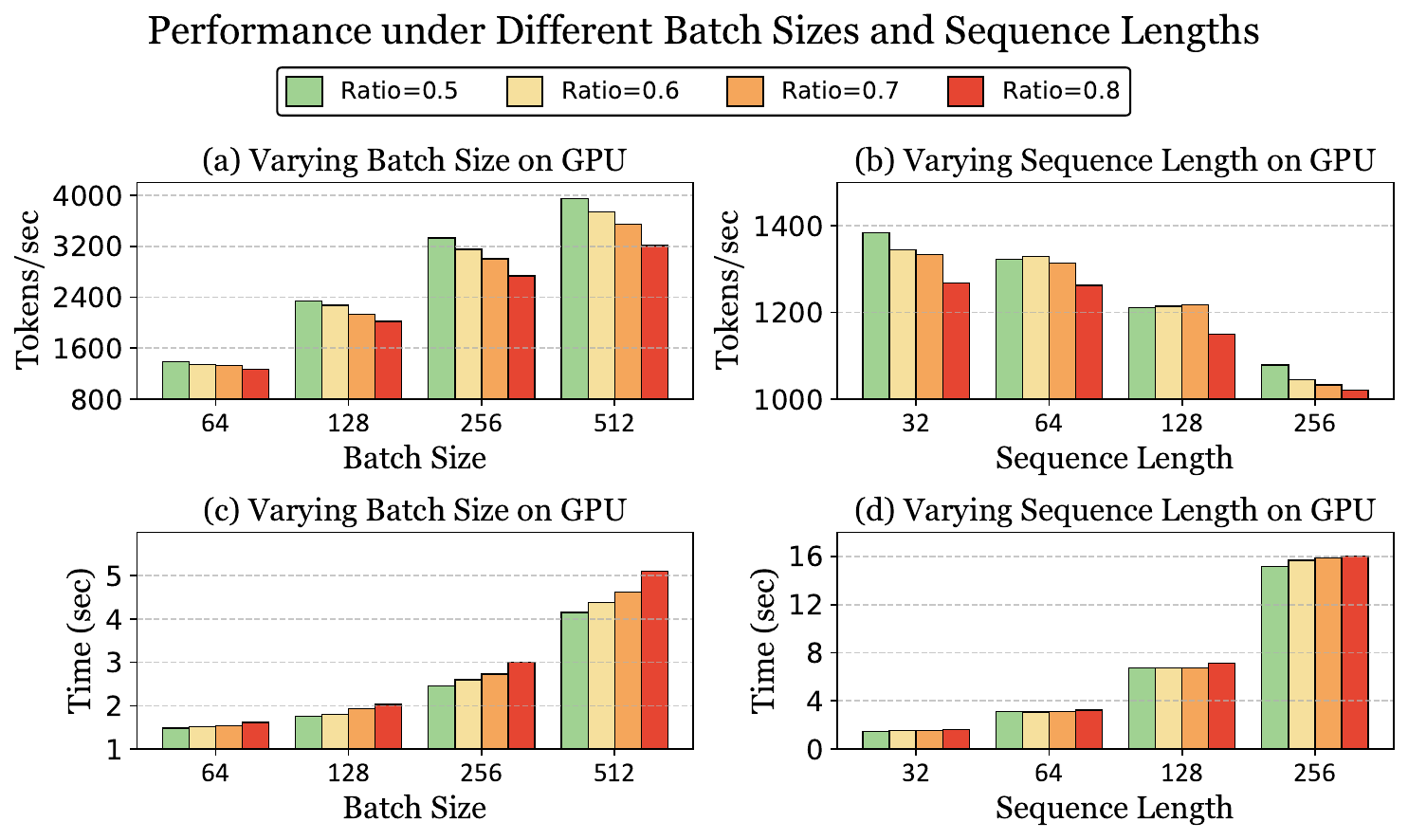}
\caption{Inference efficiency on Vicuna-7B under different batch sizes and sequence lengths.}
\label{inference}
\end{figure}

When applied to Vicuna-7B's 32-layer Transformer architecture, this approach demonstrates three key advantages: (1) The self-attention module's Q/K/V projections (original $3Ld^2$ FLOPs) reduce to $2Ldk$ operations, while the FFN layers' dense matrices ($8Ld^2$ FLOPs) compress to $4Ldk$ operations, where $d=4096$ and $k\approx 0.6d$. (2) The 40\% parameter reduction decreases memory bandwidth pressure, with model size shrinking from 14GB to 8.4GB in FP16 format. Fig.\ref{inference} presents the throughput and inference speed of Vicuna-7B model compressed by DipSVD across varying batch sizes and sequence lengths. As anticipated, higher compression ratios yield measurable improvements in both throughput and inference speed.

\section{Supplementary Experiment Results}

\subsection{Detailed performance}
Tab.\ref{appendix_label1}–\ref{appendix_label5} present a comprehensive comparison between our proposed DipSVD method and existing baselines across five foundational models. The results clearly demonstrate the effectiveness of our approach, with evaluations on three language generation datasets (measured by perplexity) and seven classification tasks (measured by accuracy). The consistent superiority of DipSVD across all benchmarks highlights its robustness and generalization capability.

\begin{table*}[ht]
\renewcommand{\arraystretch}{1.2} 
\resizebox{1.0\textwidth}{!}{%  
\begin{tabular}{cc|ccc|cccccccc} 
\hline \hline 
Radio & Method & WikiText-2 & PTB & C4 & Openb. & ARC\_e & WinoG. & HellaS. & ARC\_c & PIQA & MathQA & Average \\ \hline
\multirow{4}{*}{0.8} & ASVD & 8.759 & \textbf{12.705} & \textbf{10.833} & \textbf{0.296} & 0.671 & \textbf{0.694} & \textbf{0.505} & \textbf{0.340} & \textbf{0.750} & 0.233 & \textbf{0.498} \\
 & FWSVD & 9.280 & 14.525 & 11.880 & 0.266 & \textbf{0.679} & 0.646 & 0.472 & 0.329 & 0.742 & 0.228 & 0.480 \\
 & SVD-LLM & \textbf{7.894} & 16.848 & 16.118 & 0.258 & 0.623 & 0.652 & 0.433 & 0.310 & 0.687 & 0.234 & 0.457 \\
 & Ours & 7.949 & 15.597 & 14.074 & 0.268 & 0.633 & 0.648 & 0.454 & 0.328 & 0.710 & \textbf{0.238} & 0.468 \\ \hline
\multirow{4}{*}{0.7} & ASVD & 95.268 & 200.937 & 86.269 & 0.186 & 0.379 & 0.557 & 0.333 & 0.242 & 0.607 & 0.218 & 0.360 \\
 & FWSVD & 33.001 & 53.587 & 38.240 & 0.186 & 0.507 & 0.572 & 0.343 & 0.242 & 0.632 & 0.217 & 0.386 \\
 & SVD-LLM & 9.526 & 28.967 & 26.390 & 0.242 & 0.509 & 0.570 & 0.352 & 0.269 & 0.630 & 0.227 & 0.400 \\
 & Ours & \textbf{9.427} & \textbf{22.270} & \textbf{19.909} & \textbf{0.242} & \textbf{0.602} & \textbf{0.640} & \textbf{0.405} & \textbf{0.296} & \textbf{0.661} & \textbf{0.230} & \textbf{0.440} \\ \hline
\multirow{4}{*}{0.6} & ASVD & 9111.411 & 19425.612 & 8676.642 & 0.158 & 0.286 & 0.486 & 0.267 & 0.210 & 0.538 & 0.204 & 0.307 \\
 & FWSVD & 199.142 & 332.344 & 255.026 & 0.158 & 0.348 & 0.526 & 0.275 & 0.195 & 0.571 & 0.211 & 0.326 \\
 & SVD-LLM & 13.854 & 63.864 & 57.281 & 0.208 & 0.455 & 0.566 & 0.323 & 0.240 & 0.598 & 0.217 & 0.372 \\
 & Ours & \textbf{12.760} & \textbf{46.951} & \textbf{34.352} & \textbf{0.222} & \textbf{0.503} & \textbf{0.613} & \textbf{0.358} & \textbf{0.277} & \textbf{0.640} & \textbf{0.224} & \textbf{0.405} \\ \hline
\multirow{4}{*}{0.5} & ASVD & 37479.324 & 57294.849 & 37767.010 & 0.124 & 0.263 & 0.508 & 0.256 & 0.212 & 0.519 & 0.202 & 0.298 \\
 & FWSVD & 4622.404 & 8861.445 & 9240.634 & 0.124 & 0.277 & 0.508 & 0.260 & 0.207 & 0.535 & 0.210 & 0.303 \\
 & SVD-LLM & 26.864 & 191.380 & 153.840 & 0.160 & 0.361 & 0.540 & 0.288 & 0.207 & 0.570 & 0.216 & 0.334 \\
 & Ours & \textbf{20.983} & \textbf{116.404} & \textbf{94.796} & \textbf{0.192} & \textbf{0.382} & \textbf{0.559} & \textbf{0.306} & \textbf{0.223} & \textbf{0.590} & \textbf{0.229} & \textbf{0.354} \\ \hline \hline
\end{tabular}
}
\caption{Overall Performance of LLama-7B.} %表注
\label{appendix_label1} 
\end{table*}

\begin{table*}[ht]
\renewcommand{\arraystretch}{1.2} 
\resizebox{1.0\textwidth}{!}{
\begin{tabular}{cc|ccc|cccccccc} 
\hline \hline
Radio & Method & WikiText-2 & PTB & C4 & Openb. & ARC\_e & WinoG. & HellaS. & ARC\_c & PIQA & MathQA & Average \\ \hline
\multirow{4}{*}{0.8} & ASVD & 13.723 & 72.029 & \textbf{18.261} & 0.264 & 0.641 & 0.624 & \textbf{0.440} & \textbf{0.347} & \textbf{0.701} & 0.237 & \textbf{0.465} \\
 & FWSVD & 15.312 & 75.042 & 19.945 & 0.238 & 0.611 & \textbf{0.627} & 0.401 & 0.309 & 0.695 & 0.233 & 0.445 \\
 & SVD-LLM & \textbf{9.942} & 71.366 & 23.358 & 0.252 & 0.579 & 0.598 & 0.401 & 0.315 & 0.666 & 0.229 & 0.434 \\
 & Ours & 9.952 & \textbf{56.869} & 19.722 & \textbf{0.268} & \textbf{0.650} & 0.598 & 0.431 & 0.345 & 0.689 & \textbf{0.247} & 0.461 \\ \hline
\multirow{4}{*}{0.7} & ASVD & 91.388 & 415.615 & 136.157 & 0.158 & 0.335 & 0.503 & 0.287 & 0.208 & 0.556 & 0.205 & 0.322 \\
 & FWSVD & 43.690 & 239.318 & 64.753 & 0.172 & 0.459 & 0.545 & 0.312 & 0.224 & 0.613 & 0.221 & 0.364 \\
 & SVD-LLM & 12.416 & 124.506 & 39.528 & 0.244 & 0.506 & 0.570 & 0.353 & 0.270 & 0.629 & 0.228 & 0.400 \\
 & Ours & \textbf{12.144} & \textbf{81.089} & \textbf{28.837} & \textbf{0.248} & \textbf{0.573} & \textbf{0.597} & \textbf{0.384} & \textbf{0.293} & \textbf{0.659} & \textbf{0.232} & \textbf{0.427} \\ \hline
\multirow{4}{*}{0.6} & ASVD & 1580.427 & 3069.448 & 1735.991 & 0.126 & 0.273 & 0.524 & 0.259 & 0.224 & 0.527 & 0.215 & 0.307 \\
 & FWSVD & 347.362 & 1711.730 & 461.874 & 0.128 & 0.293 & 0.519 & 0.267 & 0.209 & 0.554 & 0.212 & 0.312 \\
 & SVD-LLM & 18.347 & 261.100 & 77.706 & 0.188 & 0.430 & 0.542 & 0.314 & 0.238 & 0.590 & 0.217 & 0.360 \\
 & Ours & \textbf{17.085} & \textbf{142.752} & \textbf{49.183} & \textbf{0.194} & \textbf{0.469} & \textbf{0.580} & \textbf{0.345} & \textbf{0.259} & \textbf{0.613} & \textbf{0.231} & \textbf{0.384} \\ \hline
\multirow{4}{*}{0.5} & ASVD & 22934.960 & 28252.915 & 24201.540 & 0.146 & 0.249 & 0.508 & 0.258 & 0.213 & 0.512 & 0.185 & 0.296 \\
 & FWSVD & 4449.084 & 9353.528 & 4421.253 & 0.128 & 0.273 & 0.501 & 0.262 & 0.210 & 0.533 & 0.201 & 0.301 \\
 & SVD-LLM & 35.569 & 615.591 & 185.780 & 0.158 & 0.338 & 0.525 & 0.286 & 0.231 & 0.567 & 0.225 & 0.333 \\
 & Ours & \textbf{27.807} & \textbf{375.093} & \textbf{111.996} & \textbf{0.170} & \textbf{0.365} & \textbf{0.543} & \textbf{0.300} & \textbf{0.220} & \textbf{0.571} & \textbf{0.221} & \textbf{0.342} \\ \hline \hline
\end{tabular}
}
\caption{Overall Performance of Vicuna-7B.} 
\label{appendix_label2}
\end{table*}

\begin{table*}[ht]
\renewcommand{\arraystretch}{1.2} 
\resizebox{1.0\textwidth}{!}{
\begin{tabular}{cc|ccc|cccccccc} 
\hline \hline
Radio & Method & WikiText-2 & PTB & C4 & Openb. & ARC\_e & WinoG. & HellaS. & ARC\_c & PIQA & MathQA & Average \\ \hline
\multirow{4}{*}{0.8} & ASVD & 12.727 & 17.215 & \textbf{16.851} & 0.288 & 0.654 & 0.650 & \textbf{0.486} & 0.325 & \textbf{0.742} & 0.237 & 0.483 \\
 & FWSVD & 15.578 & 23.550 & 24.258 & 0.238 & 0.617 & 0.643 & 0.411 & 0.274 & 0.714 & 0.242 & 0.448 \\
 & SVD-LLM & 9.128 & 19.106 & 21.004 & 0.286 & 0.636 & 0.650 & 0.431 & 0.330 & 0.711 & 0.243 & 0.470 \\
 & Ours & \textbf{8.805} & \textbf{15.890} & 17.732 & \textbf{0.290} & \textbf{0.665} & \textbf{0.663} & 0.450 & \textbf{0.342} & 0.721 & \textbf{0.256} & \textbf{0.484} \\ \hline
\multirow{4}{*}{0.7} & ASVD & 85.169 & 87.709 & 79.853 & 0.154 & 0.390 & 0.516 & 0.312 & 0.213 & 0.610 & 0.210 & 0.344 \\
 & FWSVD & 68.416 & 99.775 & 118.319 & 0.142 & 0.406 & 0.551 & 0.296 & 0.194 & 0.595 & 0.220 & 0.344 \\
 & SVD-LLM & 10.841 & 30.747 & 32.622 & 0.260 & 0.589 & 0.609 & 0.384 & 0.283 & 0.670 & 0.232 & 0.432 \\
 & Ours & \textbf{9.895} & \textbf{20.977} & \textbf{22.558} & \textbf{0.276} & \textbf{0.628} & \textbf{0.631} & \textbf{0.415} & \textbf{0.312} & \textbf{0.700} & \textbf{0.239} & \textbf{0.457} \\ \hline
\multirow{4}{*}{0.6} & ASVD & 3806.825 & 7580.528 & 4355.394 & 0.140 & 0.298 & 0.494 & 0.267 & 0.202 & 0.546 & 0.209 & 0.308 \\
 & FWSVD & 202.822 & 265.391 & 325.196 & 0.126 & 0.309 & 0.499 & 0.267 & 0.184 & 0.545 & 0.215 & 0.306 \\
 & SVD-LLM & 14.449 & 55.803 & 58.199 & 0.228 & 0.529 & 0.578 & 0.336 & 0.241 & 0.626 & 0.225 & 0.395 \\
 & Ours & \textbf{12.077} & \textbf{32.890} & \textbf{35.540} & \textbf{0.250} & \textbf{0.572} & \textbf{0.624} & \textbf{0.366} & \textbf{0.285} & \textbf{0.653} & \textbf{0.233} & \textbf{0.426} \\ \hline
\multirow{4}{*}{0.5} & ASVD & 64971.820 & 99927.992 & 57731.498 & 0.128 & 0.262 & 0.498 & 0.263 & 0.213 & 0.515 & 0.201 & 0.297 \\
 & FWSVD & 600.397 & 890.699 & 774.628 & 0.132 & 0.281 & 0.478 & 0.264 & 0.195 & 0.531 & 0.205 & 0.298 \\
 & SVD-LLM & 22.660 & 132.187 & 117.580 & 0.192 & 0.427 & 0.538 & 0.300 & 0.220 & 0.591 & 0.213 & 0.355 \\
 & Ours & \textbf{17.960} & \textbf{79.482} & \textbf{73.707} & \textbf{0.194} & \textbf{0.468} & \textbf{0.554} & \textbf{0.316} & \textbf{0.222} & \textbf{0.609} & \textbf{0.223} & \textbf{0.369} \\ \hline \hline
\end{tabular}
}
\caption{Overall Performance of Deepseek-7B.} 
\label{appendix_label3}
\end{table*}

\begin{table*}[ht]
\renewcommand{\arraystretch}{1.2} 
\resizebox{1.0\textwidth}{!}{
\begin{tabular}{cc|ccc|cccccccc} 
\hline \hline
Radio & Method & WikiText-2 & PTB & C4 & Openb. & ARC\_e & WinoG. & HellaS. & ARC\_c & PIQA & MathQA & Average \\ \hline
\multirow{4}{*}{0.8} & ASVD & 6.743 & \textbf{10.407} & \textbf{9.123} & \textbf{0.330} & 0.732 & \textbf{0.707} & 0.540 & 0.431 & 0.771 & \textbf{0.251} & \textbf{0.538} \\
 & FWSVD & 6.938 & 11.231 & 9.499 & 0.312 & \textbf{0.737} & 0.696 & 0.518 & 0.393 & 0.769 & 0.258 & 0.526 \\
 & SVD-LLM & \textbf{6.575} & 12.194 & 12.811 & 0.302 & 0.683 & 0.684 & 0.470 & 0.356 & 0.725 & 0.265 & 0.498 \\
 & Ours & 6.649 & 11.383 & 11.418 & 0.306 & 0.681 & 0.692 & 0.490 & 0.369 & 0.734 & \textbf{0.258} & 0.503 \\ \hline
\multirow{4}{*}{0.7} & ASVD & 17.648 & 32.963 & 20.866 & 0.218 & 0.551 & 0.611 & 0.398 & 0.288 & 0.690 & 0.217 & 0.425 \\
 & FWSVD & 12.963 & 22.123 & 18.509 & 0.248 & 0.632 & 0.640 & 0.403 & 0.294 & \textbf{0.707} & 0.229 & 0.450 \\
 & SVD-LLM & \textbf{7.618} & 17.823 & 18.825 & 0.276 & 0.619 & 0.672 & 0.415 & 0.300 & 0.671 & 0.242 & 0.457 \\
 & Ours & 7.697 & \textbf{15.681} & \textbf{14.614} & \textbf{0.284} & \textbf{0.641} & \textbf{0.656} & \textbf{0.449} & \textbf{0.328} & 0.694 & \textbf{0.249} & \textbf{0.472} \\ \hline
\multirow{4}{*}{0.6} & ASVD & 201.027 & 286.850 & 183.898 & 0.148 & 0.336 & 0.518 & 0.293 & 0.197 & 0.579 & 0.215 & 0.327 \\
 & FWSVD & 45.150 & 75.662 & 64.610 & 0.166 & 0.431 & 0.552 & 0.305 & 0.219 & 0.607 & 0.226 & 0.358 \\
 & SVD-LLM & 9.836 & 34.222 & 33.328 & 0.222 & 0.521 & 0.639 & 0.355 & 0.248 & 0.637 & 0.228 & 0.407 \\
 & Ours & \textbf{9.575} & \textbf{25.782} & \textbf{21.716} & \textbf{0.230} & \textbf{0.548} & \textbf{0.644} & \textbf{0.402} & \textbf{0.283} & \textbf{0.661} & \textbf{0.233} & \textbf{0.429} \\ \hline
\multirow{4}{*}{0.5} & ASVD & 11445.274 & 13304.711 & 10897.571 & 0.116 & 0.270 & 0.490 & 0.267 & 0.221 & 0.535 & 0.206 & 0.301 \\
 & FWSVD & 193.531 & 275.487 & 245.409 & 0.126 & 0.304 & 0.518 & 0.270 & 0.176 & 0.550 & 0.213 & 0.308 \\
 & SVD-LLM & 14.984 & 89.288 & 68.417 & \textbf{0.194} & 0.420 & 0.582 & 0.314 & 0.224 & 0.588 & 0.220 & 0.363 \\
 & Ours & \textbf{13.526} & \textbf{74.449} & \textbf{45.166} & 0.192 & \textbf{0.447} & \textbf{0.603} & \textbf{0.333} & \textbf{0.229} & \textbf{0.612} & \textbf{0.229} & \textbf{0.378} \\ \hline \hline
\end{tabular}
}
\caption{Overall Performance of LLama-13B.} 
\label{appendix_label4} 
\end{table*}

\begin{table*}[ht]
\renewcommand{\arraystretch}{1.2} 
\resizebox{1.0\textwidth}{!}{
\begin{tabular}{cc|ccc|cccccccc} 
\hline \hline
Radio & Method & WikiText-2 & PTB & C4 & Openb. & ARC\_e & WinoG. & HellaS. & ARC\_c & PIQA & MathQA & Average \\ \hline
\multirow{4}{*}{0.8} & ASVD & 10.261 & 135.308 & 14.523 & 0.276 & 0.684 & 0.663 & 0.466 & 0.368 & \textbf{0.739} & 0.259 & 0.493 \\
 & FWSVD & 12.636 & 152.403 & 19.540 & 0.230 & 0.657 & 0.624 & 0.412 & 0.312 & 0.717 & 0.237 & 0.456 \\
 & SVD-LLM & 8.035 & 82.811 & 18.269 & \textbf{0.314} & \textbf{0.687} & 0.671 & 0.456 & \textbf{0.378} & 0.727 & \textbf{0.264} & \textbf{0.500} \\
 & Ours & \textbf{7.892} & \textbf{36.405} & \textbf{14.225} & 0.306 & 0.666 & \textbf{0.674} & \textbf{0.471} & 0.357 & 0.718 & 0.262 & 0.493 \\ \hline
\multirow{4}{*}{0.7} & ASVD & 28.309 & 637.196 & 39.799 & 0.220 & 0.528 & 0.574 & 0.340 & 0.250 & 0.669 & 0.226 & 0.401 \\
 & FWSVD & 32.715 & 310.304 & 47.408 & 0.188 & 0.506 & 0.564 & 0.317 & 0.234 & 0.647 & 0.223 & 0.383 \\
 & SVD-LLM & 9.616 & 145.715 & 29.204 & \textbf{0.284} & \textbf{0.620} & 0.641 & 0.390 & 0.300 & 0.656 & \textbf{0.250} & 0.449 \\
 & Ours & \textbf{9.070} & \textbf{49.948} & \textbf{19.203} & 0.268 & 0.610 & \textbf{0.657} & \textbf{0.419} & \textbf{0.320} & \textbf{0.688} & 0.243 & \textbf{0.458} \\ \hline
\multirow{4}{*}{0.6} & ASVD & 189.392 & 1601.441 & 190.550 & 0.128 & 0.310 & 0.518 & 0.276 & 0.202 & 0.572 & 0.214 & 0.317 \\
 & FWSVD & 106.356 & 668.389 & 146.253 & 0.130 & 0.307 & 0.510 & 0.278 & 0.190 & 0.550 & 0.218 & 0.312 \\
 & SVD-LLM & 13.221 & 324.892 & 58.335 & \textbf{0.244} & 0.506 & 0.600 & 0.333 & 0.237 & 0.598 & \textbf{0.232} & 0.393 \\
 & Ours & \textbf{12.138} & \textbf{73.798} & \textbf{32.224} & 0.230 & \textbf{0.511} & \textbf{0.609} & \textbf{0.358} & \textbf{0.273} & \textbf{0.640} & 0.227 & \textbf{0.407} \\ \hline
\multirow{4}{*}{0.5} & ASVD & 1220.318 & 3922.871 & 1080.129 & 0.144 & 0.275 & 0.485 & 0.262 & \textbf{0.228} & 0.536 & 0.209 & 0.306 \\
 & FWSVD & 304.176 & 1475.578 & 395.776 & 0.124 & 0.284 & 0.489 & 0.264 & 0.207 & 0.540 & 0.202 & 0.301 \\
 & SVD-LLM & 23.607 & 699.598 & 140.584 & 0.172 & 0.372 & 0.563 & 0.296 & 0.218 & 0.563 & \textbf{0.220} & 0.343 \\
 & Ours & \textbf{18.150} & \textbf{148.365} & \textbf{71.652} & \textbf{0.184} & \textbf{0.404} & \textbf{0.563} & \textbf{0.311} & 0.215 & \textbf{0.582} & 0.217 & \textbf{0.354} \\ \hline \hline
\end{tabular}
}
\caption{Overall Performance of Vicuna-13B.} 
\label{appendix_label5}
\end{table*}

\subsection{Contents Generated from the compressed model}
In this section, we present sample outputs generated by our compressed LLaMA-7B model using th proposed DipSVD method at varying compression ratios. As shown in Tab.\ref{description}, our DipSVD method can maintain high-quality text generation across different compression levels, showcasing its robustness in model compression.

\begin{table*}[t]
\centering
\begin{tabular}{@{}lp{15cm}@{}}
\toprule
\textbf{Score} & \textbf{Description} \\
\midrule
\multicolumn{2}{l}{\textbf{What is the responsibility of the AI assistant?}} \\
1 & AI assistants are responsible for understanding and responding to the user’s requests. They also need to be able to make decisions and take action on behalf of the user. For example, if a user asks Alexa to turn on the lights in the living room, Alexa needs to understand what “turn on” means and then activate the light bulbs accordingly. \\
0.8 & The AI assists you in your work. For example, if you are driving a car, it helps you to find parking spaces. If you're looking for a book, it shows you where the book is located. \\
0.7 & The AI assists the player in various ways. For example, if the player has not completed a particular task, such as collecting an item or completing a puzzle, they are prompted to do so by their AI. \\
0.6 & How AI assistants can be used by AI researchers and AI scientists. The AI Assistant in AI has made AI more important as AI evolves into AI for AI. \\
0.5 & The AI assistant, known as a robot, operates autonomously in an autonomic environment. Robots can operate autonomically and have their own autonomic system. \\
\midrule
\multicolumn{2}{l}{\textbf{What is the universe? Explain this concept.}} \\
1 & The Universe is everything that exists. It includes all of the planets, stars, galaxies, and other celestial bodies in our solar system, as well as invisible elements like dark matter and dark energy. \\
0.8 & The universe is everything that is, will be, or may be. It includes all matter, energy, antimatter, stars, galaxies, planetary systems, interstellar gases, asteroids, comets, and nebulae. \\
0.7 & The universe is made up of billions of galaxies, which contain hundreds of millions of stars. Each star has its own orbit around the center of the galaxy. \\
0.6 & The universe is a collection of galaxies bound by interstellar gases. A single galaxy may contain up to 100 million stars, and hundreds of such galaxies exist. \\
0.5 & The universe may not be defined solely by cosmological principles like astronomy and cosmology. \\
\bottomrule
\end{tabular}
\caption{Sample outputs generated by DipSVD of LLaMA-7B at varing compression ratios.}
\label{description}
\end{table*}

\end{document}